\documentclass[lettersize,journal]{IEEEtran}
\usepackage{amsmath,amsfonts}

\usepackage{textcomp}
\usepackage{stfloats}
\usepackage{url}
\usepackage{verbatim}
\usepackage{graphicx}
\usepackage{cite}

\usepackage{caption}
\usepackage{tabularx}
\usepackage{times}
\usepackage{amssymb}
\usepackage{fancyhdr}
\usepackage{epsfig}
\usepackage{subfigure}
\usepackage{url}
\usepackage{color}
\usepackage{setspace}
\usepackage{latexsym}
\usepackage{array}
\usepackage{wrapfig}
\usepackage{algorithmic}
\usepackage{multirow}
\usepackage{multicol}
\usepackage{algorithm}
\usepackage[T1]{fontenc}
\usepackage{amsmath}
\usepackage{hyperref}

\usepackage[cmintegrals]{newtxmath}

\begin{document}

\title{SAMEdge: An Edge-cloud Video Analytics Architecture for the Segment Anything Model}

\author{
\IEEEauthorblockN{ 
Rui Lu\IEEEauthorrefmark{1},
Siping Shi\IEEEauthorrefmark{1},
Yanting Liu\IEEEauthorrefmark{2},
Dan Wang\IEEEauthorrefmark{1}
}
\IEEEauthorblockA{
\\ Department of Computing, The Hong Kong Polytechnic University\\Email:\IEEEauthorrefmark{1}\{csrlu, cssshi, csdwang\}@comp.polyu.edu.hk, \IEEEauthorrefmark{2}ulysses.liu@connect.polyu.hk
}
}

\maketitle
\begin{abstract}
As artificial intelligence continues to evolve, it is increasingly capable of handling a wide range of video analytics tasks with merely one large model. One of the key foundation technologies is the Segment Anything Model (SAM), which allows the video analytics tasks to be determined on the fly according to the input prompts from the user. However, achieving real-time response in video analytics applications is crucial for user experiences due to the limited communication and computation resources on the edge, especially with SAM, where users may continuously interact by adding or adjusting prompts.

In this paper, we propose SAMEdge, a novel edge-cloud computing architecture designed to support SAM computations for edge users. SAMEdge integrates new modules on the edge and the cloud to maximize analytics accuracy under visual prompts and image prompts input with latency constraints. It addresses resource challenges associated with prompt encoding and image encoding by offering a visual prompt transformation algorithm for visual prompts and efficient workload partitioning for image encoding. 
SAMEdge is implemented by extending the open-source SAM project from Meta AI. We demonstrate the practical application of SAMEdge through a case study on a Visual Tour Guide application. Our evaluation indicates that SAMEdge significantly enhances the accuracy of the video analytics application under distinct network bandwidths across various prompts.
\end{abstract}

\IEEEpeerreviewmaketitle

\begin{figure*}[th]
\centering
\begin{minipage}[t]{0.95\textwidth}
    \centering
    \subfigure[Model training]{\includegraphics[height=120pt]{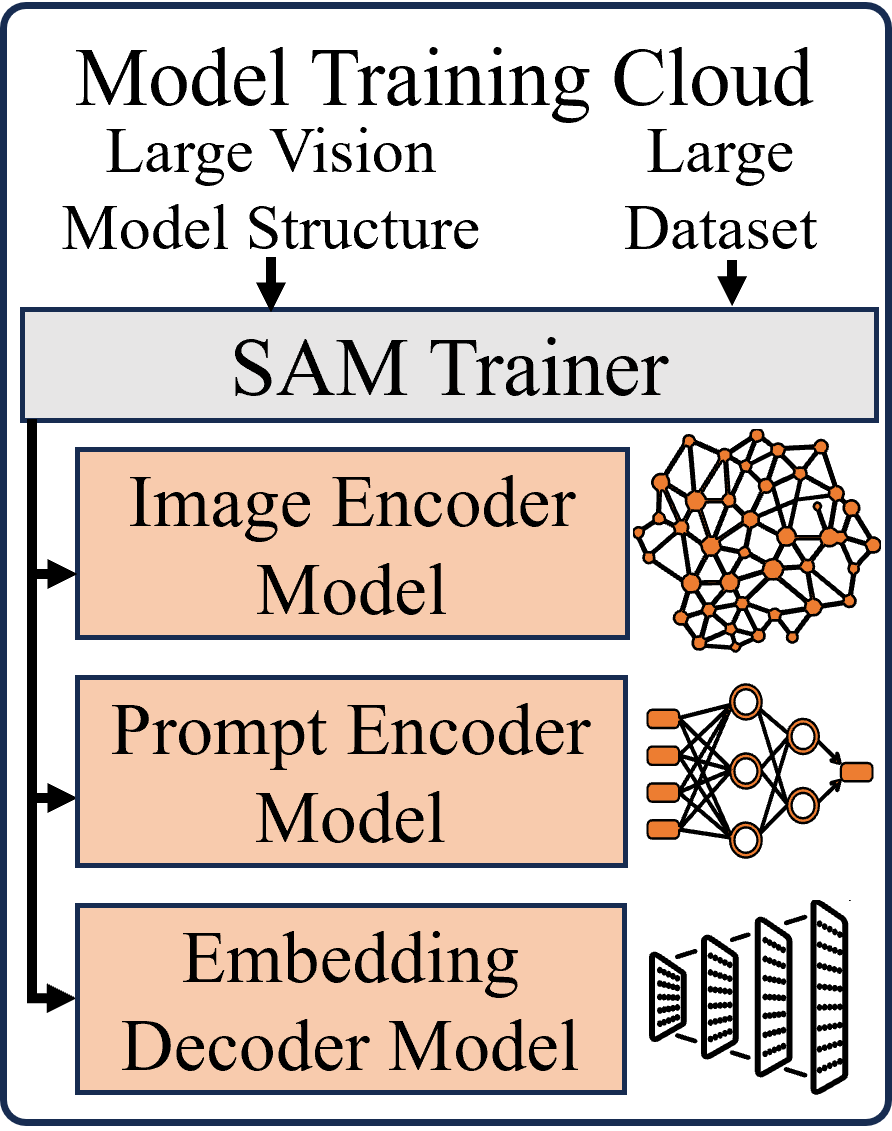}}
    \subfigure[Model inference with offline image encoding]{\includegraphics[height=120pt]{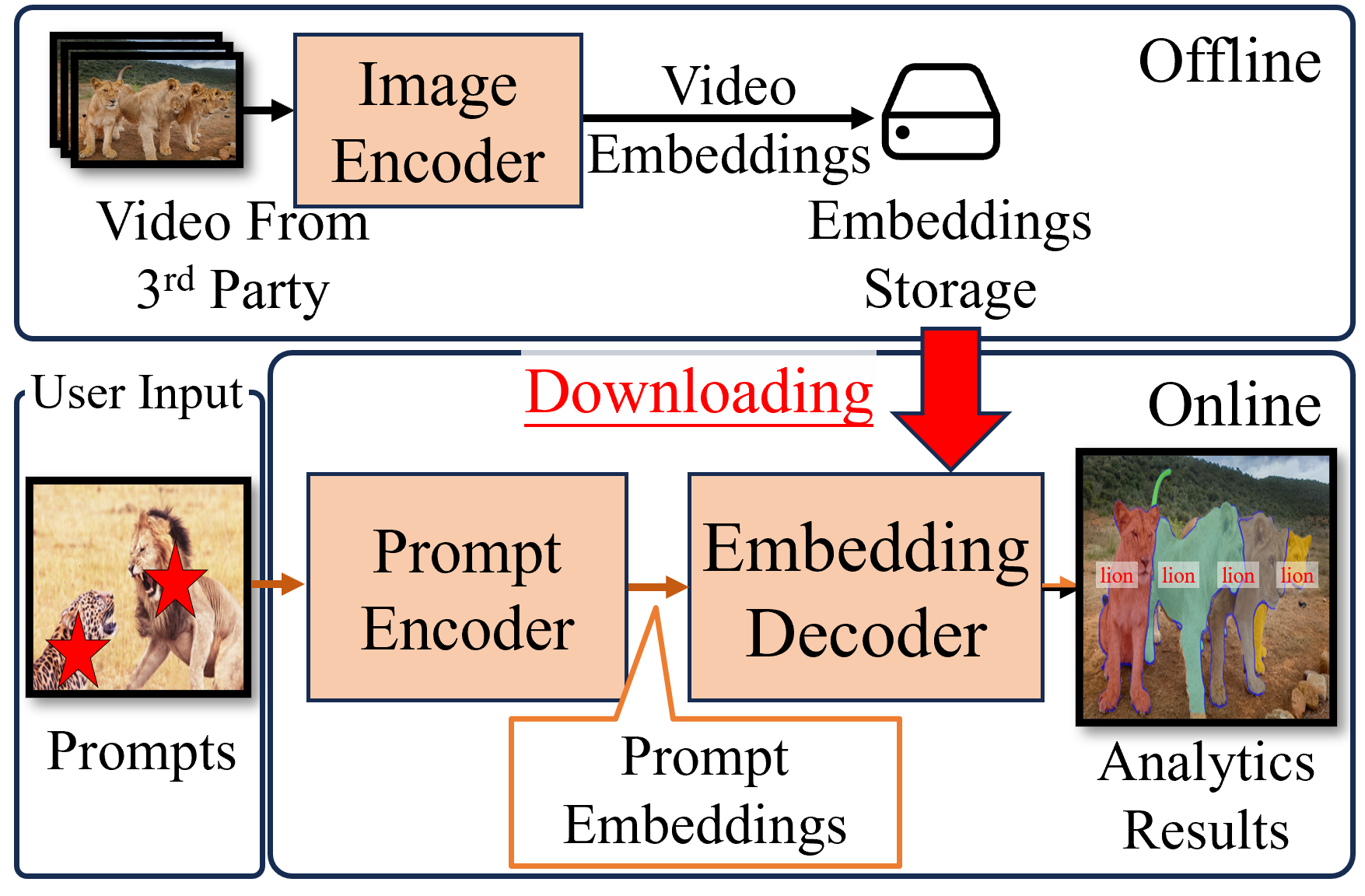}}
    \subfigure[Model inference with online image encoding]{\includegraphics[height=120pt]{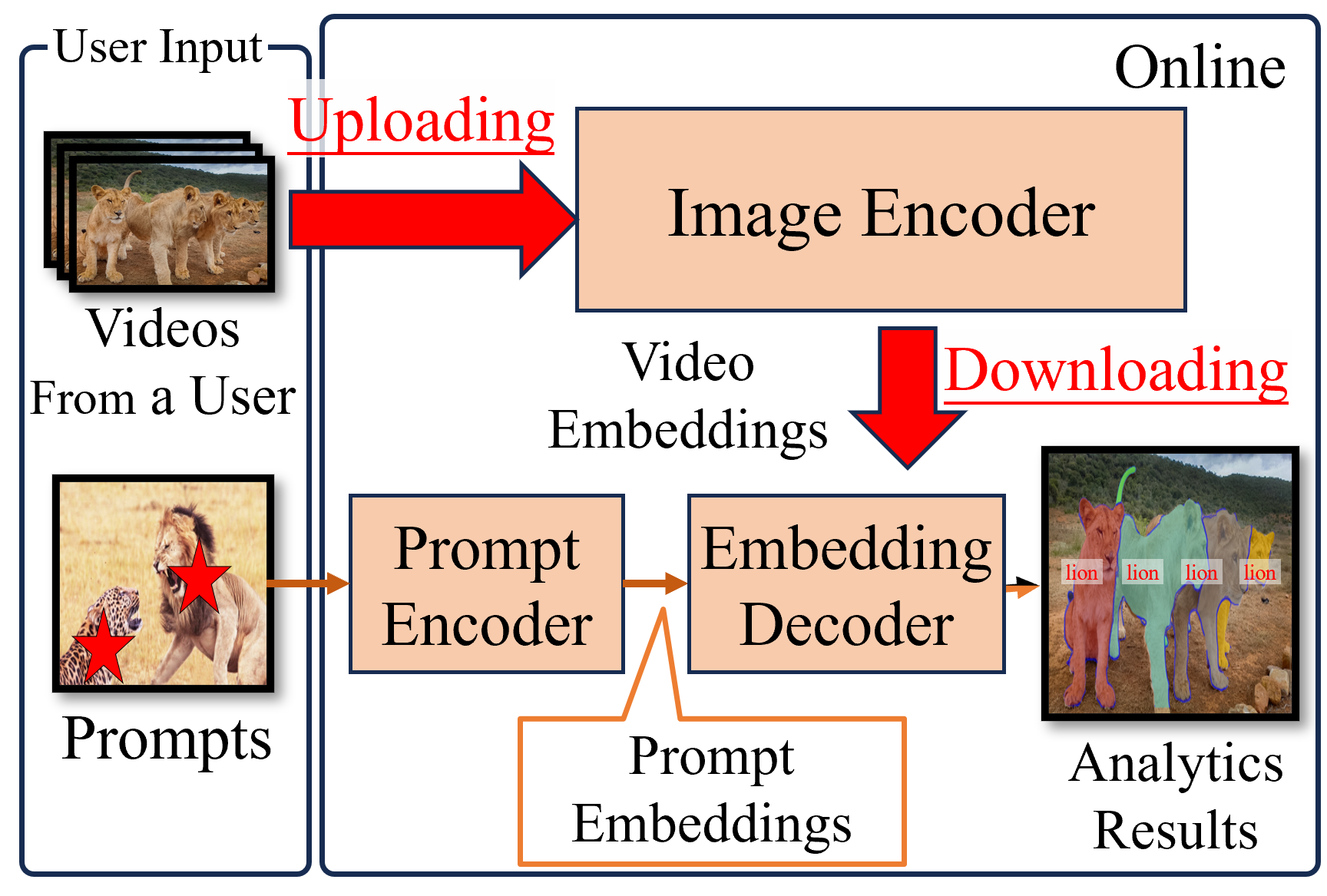}}
    \caption{The SAM computing architecture: (a) Model training, (b) Pre-recorded videos and online prompts; offline image encoding (c) Online videos and online prompts; online image encoding}
    \label{fig:sam}
\end{minipage}
\end{figure*}

\section{Introduction}\label{sec:intro}
Currently, artificial intelligence (AI) is evolving towards artificial general intelligence (AGI), which refers to the ability of an AI model to perform a wide range of tasks~\cite{al2020generalizing}. One of the key foundation technologies is the segment anything model (SAM), introduced by Meta AI Research~\cite{SAMMetaAI}, considered the "GPT Moment" in the realm of computer vision~\cite{zhang2023gptm}. It can segment any nominated objects on a certain image by giving a \textit{prompt}, i.e., the instructions of a specific downstream task. Typical prompts include \textit{visual prompts}, i.e., points, boxes, and scribbles; and \textit{image prompts}, i.e., reference images. This advanced scheme allows the video tasks to be determined on-the-fly according to the input prompts from users. 

One of the typical applications is \textit{Media Analytics Services (MAS):} A media analytics service provider allows media analytics on a video stream of interest (e.g., the BBC animal series). Users can give prompts on the tasks that they want to perform out of this video stream. For example, a user may want to find the predator animals of Africa. He can either view a few frames and input visual prompts or he can upload image prompts of predators (e.g., lions, leopards). The system will return the frames with predators from the video stream of interest. Downstream tasks such as predictor counting can also be developed. Note that in conventional systems, a specific neural network (NN) model of predator detection needs to be trained (and a set of labeled images dedicated to this task should be constructed).

Existing SAM architectures deployed on edges straightforwardly have two major challenges, including communication and computation resource constraints on edge devices. 
Communication resource consumption occurs in video and embedding transmission. Existing architectures either pre-encoding the third-party videos into embeddings on the cloud and transmit them to edge, like MAS, or users upload the video they captured to the cloud and get embedding returns. These transmissions will cost significant network bandwidth requirements. For example, the Meta AI SAM~\cite{SAMMetaAI} requires 3.7 seconds to upload a 1080p image and 1.2 seconds to download its embedding under 20 Mbps bandwidth, posing challenges for real-time response in edge user scenarios.

Computation resource consumption occurs during the encoding of prompts and images when going through \textit{prompt encoding} and \textit{image encoding} to generate prompt embeddings and video embeddings, respectively. Those embeddings are then used in \textit{embedding decoding} to generate analytics results. The new computing workloads arising are those associated with and incurred by prompts. There are new computational challenges when they are executed in the edge: (1) there can be multiple visual prompts; and decoding these prompts one-by-one leads to significant resource consumption; (2) image prompt encoding and image encoding request for workload offloading to the cloud; and these two modules can have contentions on the computing and communication resources.

In this paper, we introduce SAMEdge, a novel edge-cloud computing architecture designed to support the Segment Anything Model (SAM) computations for edge users. SAMEdge extends the capabilities of Meta AI SAM by incorporating new modules on the edge and the cloud. Our work involves the development of innovative algorithms, specifically targeting visual prompt transformation and workload partitioning. The primary objective of these algorithms is to optimize analytics accuracy with visual and image prompt input under resources and latency constraints.
To validate our approach, we implement a prototype that seamlessly integrates our research findings into the open-source SAM system developed by Meta AI. Furthermore, we showcase the practical application of SAMEdge by creating a Visual Tour Guide (VTG) application, serving as a visual co-pilot for tourists. 
Our experiments utilize Nvidia Jetson Nano as a testbed, and we conduct a comprehensive performance evaluation of SAMEdge against three comparison methods: a conventional SAM approach, a knowledge distillation SAM method, and a super-resolution method. The results reveal that SAMEdge significantly enhances video analytics accuracy, achieving up to a 4.64 times and 1.81 times improvement with visual and image prompt inputs, respectively.

In summary, the contributions of this paper are:
\begin{itemize}
    \item We develop a new edge-cloud computing architecture SAMEdge, with new modules on the edge and the cloud to support SAM computing for edge users. 
    \item We develop new algorithms, specifically, new visual prompt transformation algorithms and new workload partitioning algorithms, with the objective of optimizing the analytics accuracy of SAMEdge.
    \item We implement a prototype to integrate our research results and show how the SAMEdge architecture can be landed into practice through a case study on a visual tour guide application. Our SAMEdge system can efficiently adapt to prompt dynamics based on network changes.
\end{itemize}

\section{Related Architecture}\label{sec:related}
Video analytics systems refer to the systems that infer important events in video streams. These systems feed video frames into an NN model that is well-trained for a specific task (e.g., vehicle detection) and conduct model inferences. 
However, with the recent revolution of large foundation model technologies, a SAM architecture allows video tasks to be determined on the fly according to users' input prompts.

In a video analytics application, achieving real-time responses is crucial for user experiences, especially in SAM when users want to interact by adding or adjusting prompts. However, the limited resources on the edge make uploading all videos to the cloud impractical due to considerable bandwidth demands, especially in dynamic and complex wireless network environments. For example, transmitting a 1080p image to the cloud for image encoding and receiving embedding returns requires 4.9 seconds. 
Additionally, video transmission raises concerns about privacy leakage risks. Complete offloading SAM on edge is also challenging. For instance, SAM demands significantly more computational resources, around 10 times more than a task-specific model like YOLOv8 designed for object detection.
One viable approach is the adoption of edge-cloud video analytics systems. In this paradigm, the edge device executes partial analytics processes, offering benefits such as enhanced privacy control, real-time responsiveness, and reduced communication overhead. 

We comment that there are architectures for deploying SAM on resource-constraint edge devices. Existing architectures based on super-resolution~\cite{zhang2020sup}, NN feature selection~\cite{hu2021feva}, etc., are employed to decrease the transmission data size to accommodate the constraint and dynamic communication resources on the edge. Others, like NN model compression technologies~\cite{zhang2023mobilesam}, reduce the NN size by model quantization, model pruning, knowledge distillation, etc., to adapt to the limited edge computational resource. 
% Some researchers develop NN acceleration approaches like early exit to escape partial NN layers, memory planning for speeding up edge inference, etc. 
SAMEdge differs since it does not modify the weights or network structure of the SAM. SAMEdge dynamically offloads the workload of conventional, well-trained SAM according to the computation and communication resources of edge devices. SAMEdge is developed based on the Meta AI SAM project, the first to propose a promptable segmentation system with zero-shot generalization. SAMEdge differs from Meta AI SAM in that SAMEdge supports additional prompt categories with the new capability of workload offloading management.

\section{SAMEdge Architecture}\label{sec:design}
\subsection{SAM Workload Analysis and Partition}
To investigate resource bottlenecks in SAM, we first analyze the workload of the conventional SAM architecture. A typical \textit{SAM computing architecture} is shown in Fig. \ref{fig:sam} (e.g., the SAM project of Meta AI). Three NN models are pre-trained in a SAM system (Fig. \ref{fig:sam} (a)), an image encoder (i.e., the LVM), a prompt encoder, and an embedding decoder. Depending on whether the video stream of interest is pre-recorded or not, there are two cases: (1) pre-recorded (Fig. \ref{fig:sam} (b)); for example, the video stream comes from a third-party provider (e.g., media analytics services (MAS)): the image encoder will offline take the video stream of interest (e.g., the BBC animal series) and LVM as inputs and output an embedding of this video stream. In runtime, users input prompts to the prompt encoder, which outputs prompt embedding. The embedding decoder will take the prompt embedding and the video embedding to output analytics results; and (2) on-the-fly (Fig. \ref{fig:sam} (c)): for example, the video stream comes from a user (e.g., a video tour guide (VTG)): the user inputs both the prompts and the video stream of interest. The image encoder will online output video embedding. 

We conducted a brief analysis of the computation resource requirements within each NN model in the SAM structure. Notably, the resource requirements are approximately ten times different for the visual prompt encoder model compared to the embedding decoder model. And when image prompts and video encoding are on edge, their resource consumption dominates.
This result reveals that: 1) Each visual prompt input in SAM necessitates a single embedding decoder for decoding. Given that the embedding decoder incurs ten times more resources than the visual prompt encoder, the bottleneck in resource consumption arises from the embedding decoder. Managing prompts judiciously becomes crucial, offering potential reductions in the number of embedding decoder executions.
2) When utilizing images as prompts and capturing video to obtain video encoding at the edge, the primary contributors to resource consumption are the image encoder and image prompt encoder. It becomes necessary to implement a joint workload partition between the edge and the cloud that involves the image prompt encoder and the image encoder.

\begin{figure}[t]
    \centering
    \includegraphics[width=0.475\textwidth]{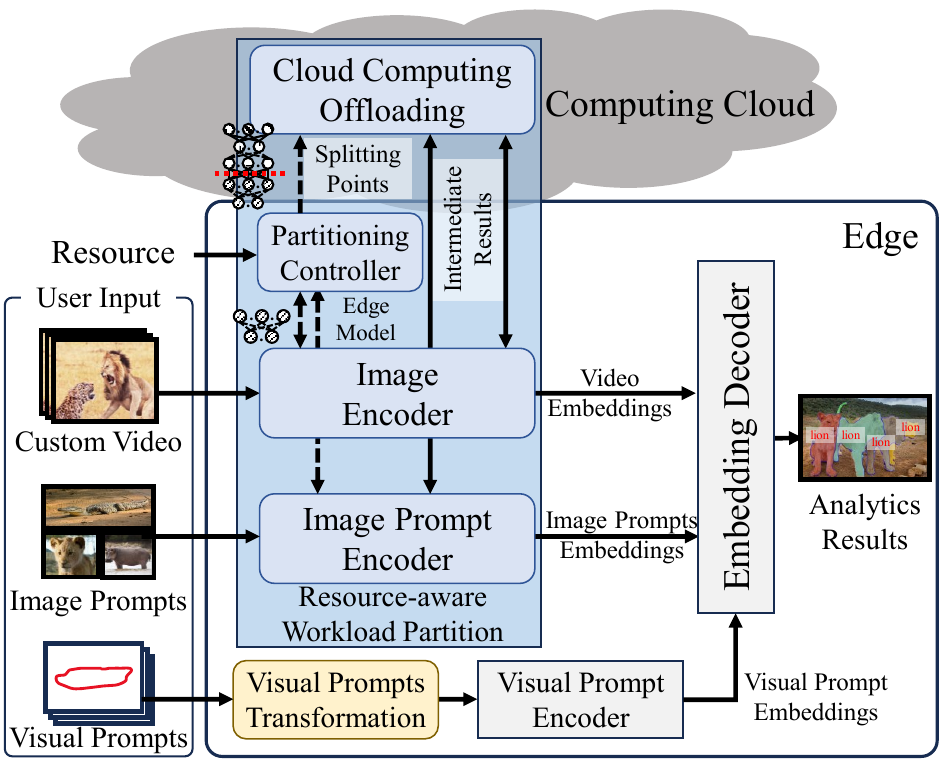}
    \caption{The architecture of SAMEdge.}
    \label{fig:framework}
\end{figure}

\subsection{The SAMEdge Architecture}
Since the conventional SAM architecture (Fig. \ref{fig:sam}) is not developed for the edge. If bringing such an architecture to the edge directly, it cannot effectively work in a real-time and resource-constrained manner. We develop a new edge-cloud computing architecture SAMEdge (Fig.~\ref{fig:framework}). SAMEdge is logically divided into the edge side and the cloud side. SAMEdge has two additional modules: 
% (1) a privacy control module for privacy-preserving image prompt encoding and image encoding; 
(1) a \textit{visual prompt transformation module} to combine and convert visual prompts; and (2) a \textit{resource-aware workload partition module} to split the image prompt encoder and image encoder.

Specifically, users initiate SAMEdge by inputting their videos and prompts, including images and visual prompts, into the edge device. Visual prompts undergo initial processing in the Visual Prompt Transformation Module, resulting in transformed visual prompts. Subsequently, the Visual Prompt Encoder refines these prompts, generating visual prompt embeddings.
Simultaneously, custom video frames and image prompts are directed to the Image Encoder and Image Prompt Encoder, respectively, inside the Resource-Aware Workload Partition Module. The partitioning controller of this module dynamically determines splitting points based on real-time edge resource availability. 
These splitting points dictate the workload division, with a portion processed locally on the edge and the remainder offloaded to a powerful computing cloud server. The encoder models will exit early at the splitting points and transmit the intermediate results to the computing cloud for further processing. 
The cloud continues executing subsequent layers and returns the computed video embeddings and image prompts embeddings to the edge. All the embedding, i.e., video embeddings, image prompts embeddings, and visual prompt embeddings, will feed into the Embedding Decoder and obtain analytics results.

\subsection{High Accuracy SAM with Communication and Computation Resource Constraints}
To design SAMEdge that supports SAM on edge devices with constraint communication and computation resources for edge users,  the objective is to maximize the video analytics accuracy under those resource and latency constraints. Specifically, we define 
\noindent\textbf{SAMEdge Resource Optimization (SERO) Problem:}
Given video frames and prompts, well-trained NN models of SAM, the computation resources at the edge and communication resources between the edge and the cloud, and a maximal latency, determine the resolutions of the video frames and the prompt images, the workload offloading on edge, and the data transmission to the cloud, so that the latency is guaranteed and the analytics accuracy is maximized.

The SERO problem exhibits a Knapsack structure, where the maximum latency constraints, layers executed on the edge, computation time on the edge, and communication time for a given SAM model can be similarly considered as the capacity of the knapsack, the weight of the item, and the value of the item, respectively. SERO is equivalent to a Knapsack problem which has been proved as NP-hard, making it impractical to find a globally optimal solution within polynomial time.

To address the SERO problem, we divided it into two subproblems based on distinct bottlenecks arising from different categories of prompts, as detailed in Section~\ref{sec:design}(A): \textit{Visual Prompts Transformation Problem} and \textit{Image Encoding Workload Partition Problem}.

\subsubsection{Visual Prompts Transformation Problem}
Prompts are assumed to be transformed into a new set with fewer prompts for conciseness through two operations: 1) combining visual prompts to reduce redundancy, and 2) converting prompt types (e.g., from points to boxes). This reduction in prompts not only decreases the number of prompts but also lowers the latency of the embedding decoder, with an impact on the precision of prompts, affecting analytics accuracy. 
So Visual Prompts Transformation Problem is defined as: 

Given a set of visual prompts, a set of visual prompt transformation operations (boxes, scribbles, etc.), and a latency constraint, determine a visual prompt transformation strategy to select and convert a new set of visual prompts, so that the latency is guaranteed and the analytics accuracy is maximized.

\textbf{Visual Prompts Transformation Algorithm}:
We propose an information-theoretical approach to address this problem. Intuitively, different prompts contain different amounts of information on the user's "instructions", which can be redundant, urging us to find a way to extract them maximally. 
Our method utilizes information entropy, a measure of information in a message, to assess NN performance, which has been applied in areas such as Explainable AI and Transfer Learning.
We design an information-based metric, \textit{visual prompt contribution score}, to measure the contribution of each visual prompt to the task, using mutual information.
More information in a prompt leads to better analytics accuracy. Let $X$ be the input prompt, and $Y$ be the predicted output for the task. The visual prompt contribution score can be represented as
\begin{equation}
C(X,Y)=I(X; Y),
\end{equation}
where $I(X;Y)=H(X)+H(Y)-H(X,Y)$ is the mutual information between $X$ and $Y$, and $H$ is the entropy \cite{huang2022entropy}.

The visual prompt contribution score forms the basis for transforming visual prompts, i.e., combining or converting them. With numerous prompt transformation options, the strategy search space is extensive.
To balance the exploration and exploitation, we develop an offline configuration and online adaption approach: 
In the offline stage, we first apply a random prompt transformation to search for the optimal transformation strategy. We then configure the mapping profiles among the transformation strategies, the analytics accuracy, and the latency. 
During the online stage, we employ a greedy algorithm. This algorithm selects the transformation strategy with the highest ratio of visual prompt contribution scores, ensuring it adheres to the latency constraint.

\subsubsection{Image Encoding Workload Partition Problem}
Given video frames and image prompts, the image encoder and the image prompt encoder NN models, the computation and communication resources at the edge, and a maximum latency constraint, determine the splitting points of the image encoder and the image prompt encoder NN models, the resolutions of the video frames and the prompt images, so that the latency is guaranteed and the analytics accuracy is maximized.

\textbf{Image Encoding Workload Partition Algorithm:}
We observe that the video embedding and the image prompt embedding output by the image encoder and the image prompt encoder should be fed into the embedding decoder at the same time for the decoding process.
As the computation of the NN models is progressive, we first construct a directed acyclic graph (DAG) where the vertex represents one NN layer.
Each layer is inherently indivisible and requires processing either on the edge side or the cloud side. To encapsulate the communication and dependency relationships among layers, a virtual entry vertex and an exit vertex are added, symbolizing the starting and ending points of the DNN, respectively.

We then develop a model partition algorithm on this DAG, where we consider the computation and communication resources and resolutions of images. 
Specifically, our objective is to identify a set of vertices, which constitutes a subset of DNN layers. The removal of this set results in the remaining graph splitting into two disconnected components. This problem can be transformed into a \textit{minimum weighted s-t cut} problem~\cite{hu2019dynamic}. Since researchers have extensively explored this problem, numerous existing solutions are available. We here employ the Boykov algorithm~\cite{boykov2004experimental} to address the model partition problem with the computational complexity of $O((m + n)n^2)$.

\section{A Case Study on Visual Tour Guide}\label{sec:imp}

\begin{figure}[t]
    \centering
    \includegraphics[width=0.475\textwidth]{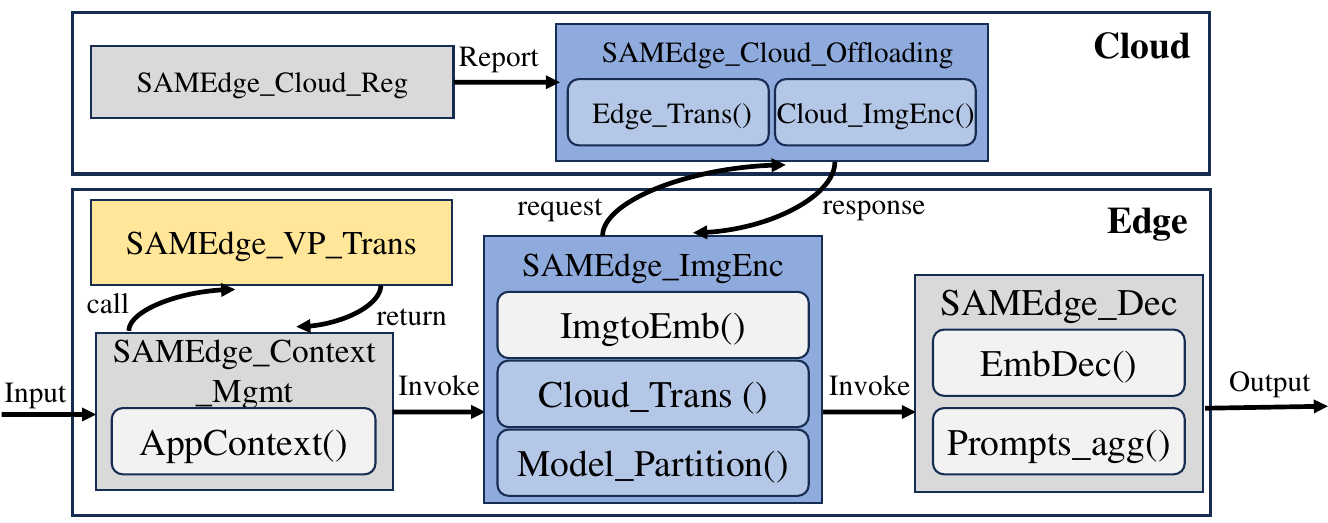}
    \caption{SAMEdge implementation based on SAM.}
    \label{fig:implete}
    \vspace{-1em}
\end{figure}

\begin{figure*}[th]
\begin{minipage}[t]{0.475\textwidth}
    \centering
\includegraphics[width=1\textwidth]{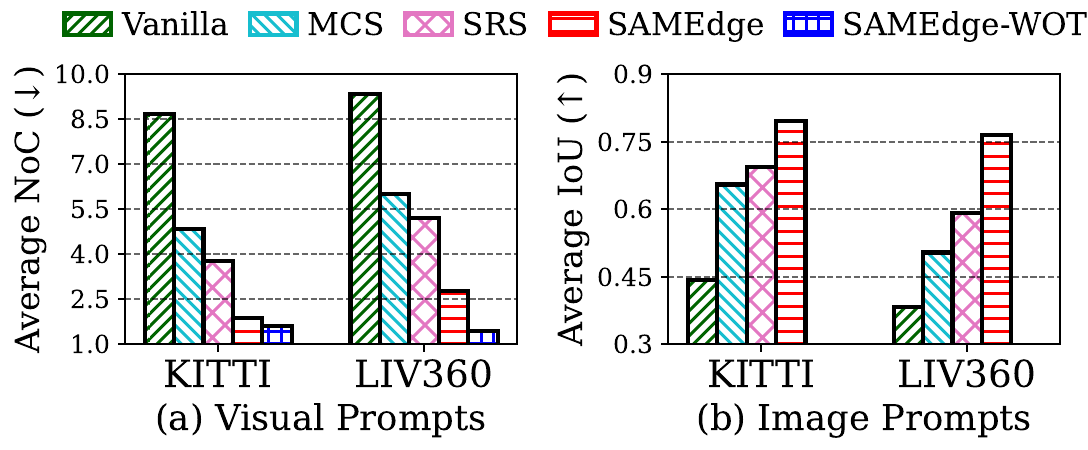}
\caption{Comparison of analytics accuracy performance under different datasets.}
    \label{fig:util}
\end{minipage}
\hfill
\begin{minipage}[t]{0.475\textwidth}
\centering
\includegraphics[width=1\textwidth]{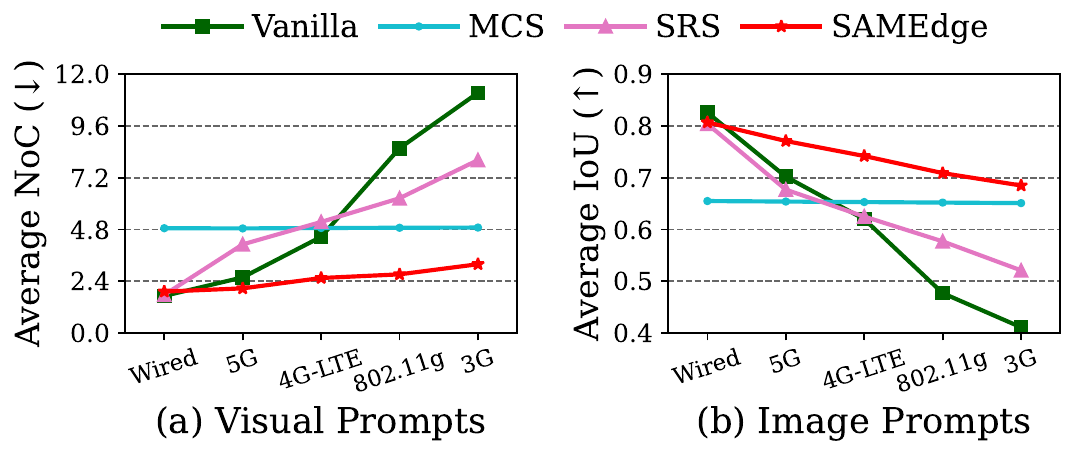}
\caption{Comparison of analytics accuracy impacted by diverse network bandwidth on KITTI.}
    \label{fig:lat}
\end{minipage}
\vspace{-1em}
\end{figure*}

\subsection{Implementation}
\subsubsection{Prototype implementation:} We implement a SAMEdge prototype by extending from Meta AI SAM~\cite{SAMMetaAI}. 
% The codes plan to be publicly available on the GitHub site.

\textbf{Briefing of the Meta SAM Architecture.} Conventional SAM has three major modules: (1) \texttt{AppContext} for prompt encoding, (2) \texttt{ImgtoEmb}, which has an image encoder model for video embedding, (3) \texttt{EmbDec}, which aggregates the prompt embedding and video embedding through an embedding decoder to output analytics results. The Meta AI SAM presumably runs in the cloud. 

\textbf{SAMEdge Module Implementation.} We develop SAMEdge, an edge-cloud SAM, by extending SAM with four modules on the edge and two on the cloud: 
(1) a \texttt{SAMEdge\_Context\_Mgmt} module that inherits \texttt{AppContext}; 
(2) a new \texttt{SAMEdge\_VP\_Trans} module for visual prompt transformation; 
(3) a \texttt{SAMEdge\_ImgEnc} module that inherits \texttt{ImgtoEmb} with a new \texttt{Model\_Partition} function to allow an early-exit of the encoder NN model at the split layer; 
(4) a \texttt{SAMEdge\_Dec} module that inherits \texttt{EmbDec}; 
(5) a \texttt{SAMEdge\_Cloud\_Reg} module to register the edge; 
and (6) a \texttt{SAMEdge\_Cloud\_Offloading} module to perform SAM inference in the cloud.

\subsubsection{VTG Implementation}
We implement VTG on SAMEdge. We use pre-train SAM of Meta AI and a TinyViT model~\cite{wu2022tinyvit} to serve as an image prompt encoder. Models are pre-trained by SA-1B~\cite{SAMMetaAI}. We fine-tune them by the street view dataset KITTI~\cite{Menze2015CVPRKITTI} and LIV360~\cite{palmer2021deep}.
In a VTG application, the prompts can change according to the user preferences and the environment. There are prompt design and engineering \cite{liu2023optimizing} on how prompts can accurately reflect user preferences. In this paper, we assume that there is a set of prompts~\cite{zou2023seem} of users for evaluation. The system adapts to prompt changes and dynamically adjusts resource consumption on edges.

\subsection{Evaluations}
\subsubsection{Evaluation Setup} 
We evaluate the performance of SAMEdge VTG  with an Nvidia Jetson Nano, a widely used edge AI device with a 128-core Maxwell GPU and a Quad-core ARM A57 CPU running the Ubuntu system. 
A workstation as the cloud server equips dual powerful NVIDIA RTX 4090 GPUs, and an Intel i9 CPU, providing exceptional NN computation capabilities.
We use two street-view benchmark datasets for model fine-tuning and inference: 
1) KITTI~\cite{Menze2015CVPRKITTI}, contains over 32,000 images of street views, 
and 2) LIV360~\cite{palmer2021deep}, contains more than 30,000 frames of 10,000 labeled objects.

\noindent\textbf{Evaluation Criteria}: We evaluate the accuracy and the delay performance of the video analytic application supported by SAMEdge. 
For image prompts, we use Intersection over Union (IoU) to measure the accuracy of VTG performance, i.e., $IoU = \textit{Area of Overlap}/\textit{Area of Union}$.
For visual prompts, we use the Number of Clicks (NoC)~\cite{zou2023seem} metric to evaluate interactive VTG performance, which measures the number of clicks of visual prompts needed to achieve a certain Intersection over Union (IoU), i.e., 90
\%. The lower value of NoC or higher value of IoU represents the higher accuracy.

\noindent\textbf{Baselines for Comparison}:
We compare SAMEdge with three existing architectures for comparison.
\begin{itemize}
    \item Vanilla only processes visual prompts and decoders on edge devices, where the image prompts and video frames are sent to the workstation for image encoding as the conventional Meta Segment Anything Model Project~\cite{SAMMetaAI}.
    \item Model Compression Scheme (MCS)
    applies a lightweight version of SAM, Mobile\_SAM~\cite{zhang2023mobilesam}, through knowledge distillation. It only has 1/60 parameters of the original SAM that can deploy on edge devices directly and ignore the influence of dynamic bandwidth.
    \item Super-resolution Scheme (SRS) \cite{zhang2020sup}
    applies a Super-resolution approach on the cloud and reduces the image resolution during transmission from the edge to the cloud according to the bandwidth. The cloud will convert the low-resolution images into high ones through the super-resolution approach for encoding.
\end{itemize}

\begin{figure*}[t]
\centering
\subfigure[Visual Prompts]{\includegraphics[width=0.365\textwidth]{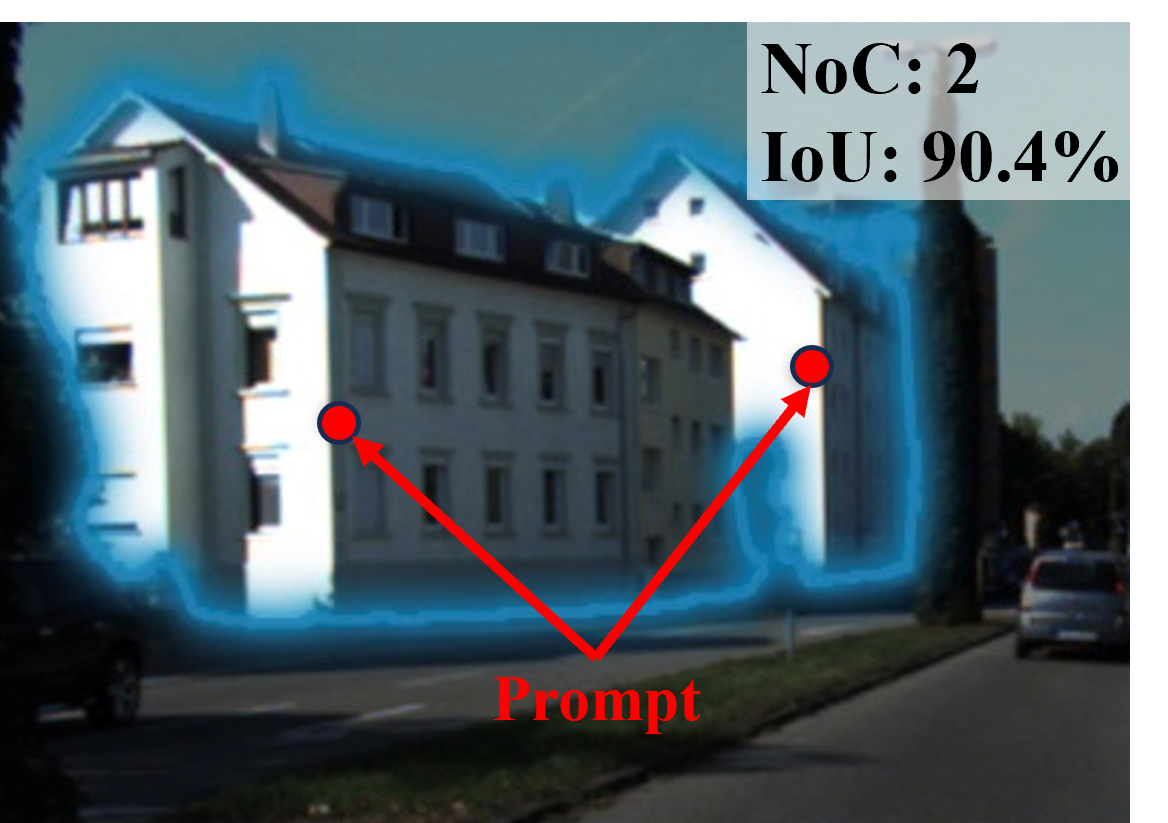}}
\hspace{2.5em}
\subfigure[Image Prompts]{\includegraphics[width=0.375\textwidth]{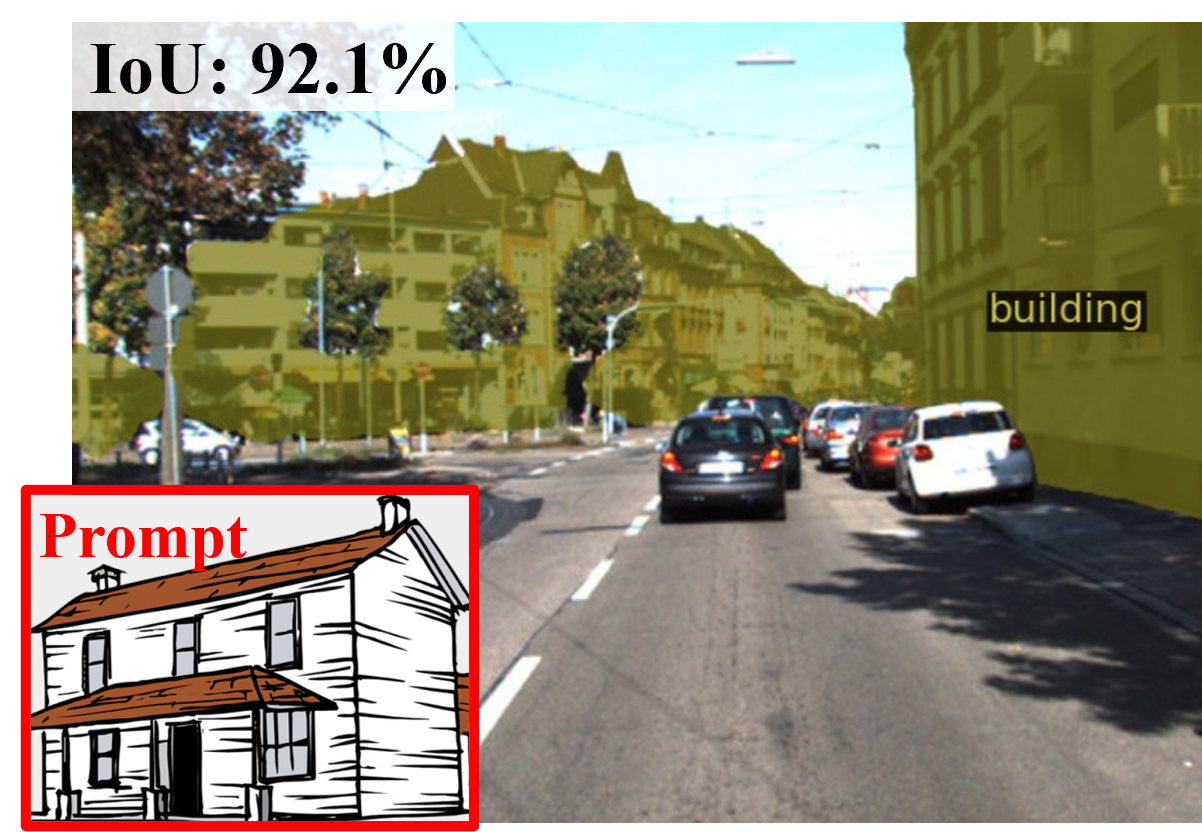}}
\caption{Visualization of SAMEdge under visual prompt and image prompt in KITTI dataset of VTG Application.}
    \label{fig:vis}
\vspace{-1em}
\end{figure*}

\subsection{Experiment Results}
\subsubsection{Improvement of Analytics Accuracy}
We evaluate the analytics accuracy performance of SAMEdge, compared to baselines on two datasets. 
Fig.~\ref{fig:util} illustrates the accuracy of (a) visual prompts and (b) image prompts measured by NoC and IoU, respectively.
Taking the KITTI dataset as an example, SAMEdge achieves 1.87 NoC and 0.797 IoU, outperforming Vanilla with 4.64 times NoC and 1.81 times IoU. This improvement stems from the fact that Vanilla transmits low-quality videos and prompts to the cloud to reduce latency, thereby compromising analytics accuracy.
MCS achieves 4.85 NoC and 0.655 IoU, exhibiting 2.59 times higher NoC and 17.8\% IoU reduction compared to SAMEdge. This reduction is attributed to model compression techniques, which not only reduce computation requirements but also degrade analytics accuracy performance.
SRS achieves 3.78 NoC and 0.694 IoU. This is because SRS reduces transmission data by downsampling resolution and then recovers it using super-resolution. However, the loss in super-resolution recovery results in significant accuracy drops. Similar results are observed in LIV360.

We further analyze the analytics performance of SAMEdge and SAMEdge-WOT in LIV360, where SAMEdge-WOT does not apply the visual transformation algorithm. SAMEdge achieves 2.77 NoC in visual prompts, while SAMEdge-WOT achieves a better value of 1.44. This difference arises because SAMEdge applies visual prompt transformation algorithms and combines prompts to reduce latency, about 68\%, with the cost of accuracy degradation.

\subsubsection{Improvement of Analytics Accuracy Impacted by Network Bandwidth}
To investigate the analytics performance across different network bandwidths, we simulate five network bandwidth traces \cite{wang2012characterizing}, including \textit{wired cabled}, \textit{5G}, \textit{4G-LTE}, \textit{802.11g}, and \textit{3G}, on KITTI. The average bandwidths are about \{1000, 100, 50, 20, 3\}Mbps, respectively. Baselines never exceed the maximum latency through their adjustable mechanisms, e.g., decline image resolutions.

The results are depicted in Fig.~\ref{fig:lat}. When the baselines and SAMEdge are provided with sufficient network bandwidth, such as with wired cabled connections, all achieve relatively high analytics accuracy under both visual and image prompts except MCS. MCS consistently performs regardless of changes in bandwidth, as it computes entirely locally, disregarding dynamic bandwidth variations. Vanilla exhibits rapid degradation with decreasing bandwidth, declining from 1.72 to 11.4 on NoC and from 0.826 to 0.411 on IoU. SRS experiences a slower decline in analytics compared to Vanilla, achieving 8.02 NoC and 0.521 IoU under the lowest bandwidth. This is because SRS reduces communication resource consumption when uploading images and videos to the cloud but still loses information during downsampling on the edge, which is challenging to recover via super-resolution on the cloud. In contrast, SAMEdge maintains high analytics accuracy across all bandwidths, experiencing only a +1.27 NoC and -0.112 IoU degradation. This is because SAMEdge only transforms embeddings to the cloud for partial inference instead of all videos and images, requiring minimal bandwidth resources. Furthermore, SAMEdge addresses the SERO problem by dynamically selecting the splitting position of the NN model constraint to maximum latency.

\subsubsection{Case study of SAMEdge on the KITTI dataset.}
We display the visualization of SAMEdge on dataset KITTI as shown in Fig.~\ref{fig:vis}. In Fig.~\ref{fig:vis}(a), the visual prompts are points clicked by the user to choose the "building"  target. Only two points are needed to mark the target with 90.4\% accuracy. In Fig.~\ref{fig:vis}(b), we input a "building" cartoon image as a reference to find similar targets, and targets are marked with 92.1\% accuracy. These validate that our proposed architecture and algorithm of SAMEdge are feasible and effective in practice.

\section{Conclusion}
In this article, we proposed SAMEdge, a novel edge-cloud computing architecture designed to support SAM computations for edge users under resources and latency constraints. 
SAMEdge fundamentally divides the workload of the conventional SAM into the edge and the cloud. We model the SERO problem and present a prompt transformation algorithm and a workload partitioning algorithm designed to maximize overall analytics accuracy within the constraints of computation and communication resources under the maximal latency on the edge device.
We implement a SAMEdge prototype that integrates Meta AI and features a Visual Tour Guide application. The results demonstrate that SAMEdge significantly enhances video analytics accuracy under dynamic network bandwidth.

\section*{BIOGRAPHIES}

\textbf{Rui Lu} received the B.S. degree from the Southern University of Science and Technology in 2019. He is currently a Ph.D. candidate at the Department of Computing of The Hong Kong Polytechnic University. His research interests include edge computing and video analytics.

\textbf{Siping Shi} received the Ph.D. degree in computer science from The Hong Kong Polytechnic University in 2023. She is currently a Postdoc at the Department of Computing at the Hong Kong Polytechnic University. Her research interests include federated learning and analytics, edge computing, and privacy-preserving machine learning systems.

\textbf{Yanting Liu} received the B.S. degree from The Hong Kong Polytechnic University in 2022. He is currently a Ph.D. student at the Department of Computing of The Hong Kong Polytechnic University. His research interests include networking, video streaming and analytics.

\textbf{Dan Wang} received the Ph.D. degree in computer science from Simon Fraser University, Canada in 2007. He is a Professor of Department of Computing, The Hong Kong Polytechnic University. He was a TPC co-Chair of IEEE/ACM IWQoS 2020 and a TPC co-Chair of ACM e-Energy 2020. He is a senior member of the IEEE.

\bibliographystyle{ieeetr}
\bibliography{ref}
\end{document}